\documentclass{article}
\usepackage[preprint]{log_2022}

\usepackage{amsmath,amssymb,amsfonts}
\usepackage{algorithmic}
\usepackage{graphicx}
\usepackage{textcomp}
\usepackage{balance}
\usepackage{array}
\usepackage{textcomp}
\usepackage{stfloats}
\usepackage{subcaption}
\usepackage{url}
\usepackage{verbatim}
\usepackage{graphicx}
\usepackage{cite}
\usepackage{times}
\usepackage{booktabs}
\usepackage{multicol}
\usepackage{multirow}
\usepackage{acronym}
\usepackage{nicematrix}
\usepackage{placeins}
\usepackage[normalem]{ulem}
\useunder{\uline}{\ul}{}
\acrodef{GNN}[GNN]{\emph{Graph Neural Network}}
\acrodefplural{GNN}[GNNs]{\emph{Graph Neural Networks}}
\acrodef{DNN}[DNN]{\emph{Deep Neural Network}}
\acrodefplural{DNN}[DNNs]{\emph{Deep Neural Networks}}
\acrodef{RNN}[RNN]{\emph{Recurrent Neural Network}}
\acrodefplural{RNN}[RNNs]{\emph{Recurrent Neural Networks}}
\acrodef{GCN}[GCN]{\emph{Graph Convolutional Network}}
\acrodefplural{GCN}[GCNs]{\emph{Graph Convolutional Networks}}
\acrodef{CNN}[CNN]{\emph{Convolutional Neural Network}}
\acrodef{FGM}[FGM]{\emph{Factorized Graph Matching}}
\acrodef{NGM-net}[NGM-net]{\emph{Neural Graph Matching Network}}
\acrodef{SGM-net}[SGM-net]{\emph{Shape Graph Matching Network}}
\acrodef{NGM-net}[NGM]{\emph{Neural Graph Matching Network}}
\acrodef{GATConv}[GATConv]{\emph{Graph Attention Networks}}

\newcommand{\real}{\ensuremath{\mathbb{R}}}

\def\etal{et al.}

\usepackage[numbers,compress,sort]{natbib}

\title{Shape-Graph Matching Network (SGM-net): Registration for Statistical Shape Analysis}

\author[Y. Zhu et al.]{%
Yanqiao Zhu\thanks{Equal contribution.}\\
\institute{University of California, Los Angeles}\\
\email{yzhu@cs.ucla.edu}\And
Yuanqi Du\footnotemark[1]\\
\institute{Cornell University}\\
\email{yd392@cornell.edu}
}

\author[]{Shenyuan Liang \\
\institute{Florida State University} \\
\email{sl20fu@fsu.edu} \And
Mauricio Pamplona Segundo \\
\institute{University of South Florida} \\
\email{mauriciop@mail.usf.edu} \And
Sathyanarayanan N. Aakur \\
\institute{Oklahoma State University} \\
\email{saakurn@okstate.edu} \And
Sudeep Sarkar \\
\institute{University of South Florida} \\
\email{sarkar@usf.edu} \And
Anuj Srivastava \\
\institute{Florida State University} \\
\email{anuj@stat.fsu.edu}
}

\begin{document}

\maketitle

\begin{abstract}
    This paper focuses on the statistical analysis of shapes of data objects called {\bf shape graphs}, a set of nodes connected by articulated curves with arbitrary shapes. A critical need here is a constrained registration of points (nodes to nodes, edges to edges) across objects. This, in turn, requires optimization over the permutation group, made challenging by differences in nodes (in terms of numbers, locations) and edges (in terms of shapes, placements, and sizes) across objects. This paper tackles this registration problem using a novel neural-network architecture and involves an unsupervised loss function developed using the {\bf elastic shape metric} for curves. This architecture results in (1) state-of-the-art matching performance and (2) an {\bf order of magnitude reduction} in the computational cost relative to baseline approaches. We demonstrate the effectiveness of the proposed approach using both simulated data and real-world 2D and 3D shape graphs. Code and data will be made publicly available after review to foster research.
\end{abstract}

\section{Introduction}
\label{sec:intro}

Using shapes to characterize objects in images and videos has wide applications in tasks such as object detection, tracking, and recognition. Given shape variability across objects and measurement errors, a {\bf statistical} analysis of shapes becomes important. There exists advanced literature~\cite{zheng2017statistical, srivastava2005statistical, dai2020statistical, kendall1977diffusion, su2020shape} on developing shape metrics, visualizing shape deformation/geodesics, computing shape summaries, building statistical shape models, and conducting shape testing for several types of objects.
A key underlying assumption in some of these approaches is that objects can be  represented by a finite number of points, called {\it landmarks}, and that the registration of landmarks, i.e., establishing dense correspondence between points across objects, is given. 
However, this assumption does not hold in general. In fact, {\bf the most challenging part of shape analysis is registration or matching, especially under a shape metric}.
Successful techniques for registration are critical for statistical shape analysis.

\begin{figure}
\centering
\includegraphics[width=.56\linewidth]{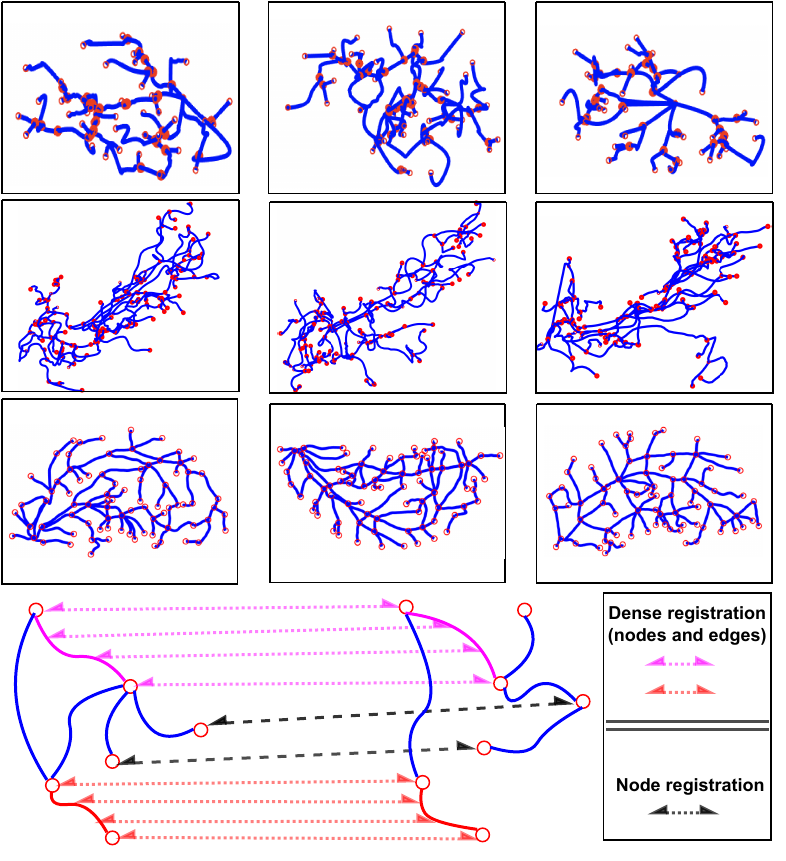}
\caption{Top:  Examples of 3D Microglia dataset. Second:  Examples of 3D brain arterial networks. Third: Examples of 2D blood vessel networks. Bottom: Registration of shape graphs requires establishing dense correspondence between nodes and points on edges across objects.}
\label{fig:examples_objects}
\end{figure}

Current shape methods, especially those relying on elastic Riemannian metrics, have addressed this need by incorporating registration as integral to shape representation. Such techniques have been developed for several types of objects, from scalar functions and Euclidean curves~\cite{FDA,younes-distance,younes-michor-mumford-shah:08} to 2D surfaces~\cite{su2020shape,jermyn2017elastic}, branching trees~\cite{duncan2018statistical,laga-roots:2019,guan-trees-arxiv:2021}, networks and graphs~\cite{guo2021quotient,jain2011graph,calissano2020populations,Kolaczyk2020AveragesOU}, and shape graphs~\cite{BAN}. 
Naturally, the registration problem gets more challenging as the objects get more complex in terms of their branching, connectivity, and articulation. For Euclidean curves, the registration corresponds to optimal re-parameterization of curves, and there are several elegant numerical solutions: dynamic programming algorithm~\cite{FDA}, gradient-based optimization~\cite{huang-etal:2015}, etc. For 2D surfaces, one can use gradient-based techniques~\cite{laga-SRNV-PAMI:2017,jermyn2017elastic} to register points across objects. For more complex objects with multiple parts or substructures, the problem becomes hierarchical, i.e., matching parts to parts and then corresponding points to points.

In this paper, we are interested in {\bf the problem of registering shape graphs under the elastic shape metric}. 
A shape graph is a collection of nodes (points in $\real^2$ or $\real^3$) connected by Euclidean curves of arbitrary shape and sizes (termed edges).
The first three rows of Fig.~\ref{fig:examples_objects} show examples of shape graphs associated with 3D Microglia, 3D brain arterial, and 2D blood vessel networks. 
Other examples include road- and river-networks, neuronal networks, and cellular structures, to name a few. 
Guo \etal~\cite{guo2021quotient,BAN} introduced a mathematical framework for statistical analysis of shape graphs, including techniques for geodesics, statistical summaries, and PCA. However, this framework has an important bottleneck. To compare the shapes of two objects, one needs to take into account their node connectivity patterns, note attributes, and the presence and shapes of edges. This comparison requires the registration of nodes and edges across shape graphs. 
Hence, registration was formulated as an optimization over the group of permutation matrices with affinity matrices derived from shape quantifications.  This results in quadratic assignment problems (QAP)~\cite{lawler1963quadratic,gold1996graduated,vogelstein2015fast,guo2021quotient} that are difficult to solve in polynomial time.  

The shape-graph registration in Guo \etal\cite{BAN} uses the Factorized Graph Matching (FGM)~\cite{fgm}, which provides good results but has a significant limitation: it is computationally expensive. 
This paper seeks a deep-learning-based solution that is an order of magnitude faster but retains the performance level. Additionally, if successful, deep learning techniques can handle large amounts of data without supervision. 
Layered structures of deep networks allow models to learn natural variations in complex data, helping solve difficult optimization problems. 
The existing literature contains several papers on using deep networks for graph match, but they focus mainly on regular (Euclidean) graphs rather than shape graphs. They study nodal registration when the node and edge attributes are Euclidean quantities and cannot directly handle {\it shapes} as edge attributes. Shapes are typically non-Euclidean variables, requiring invariance to rigid motions, and cannot directly be input in neural works designed for Euclidean data. 

The \textbf{contributions} of this paper are three-fold: (i) We solve the problem of shape graph registration using an \textit{unsupervised}, deep learning framework called \ac{SGM-net}. It is, to the best of our knowledge, the first learning-based technique to solve the most general QAP form -- Lawler's QAP~\cite{lawler1963quadratic} using factorization~\cite{fgm}, while avoiding computation of a large $O(n^4)$ affinity matrix; (ii) We demonstrate that the use of hypothetical null nodes can handle graph registration between graphs of different (node and edge) sizes. In this context, matching real nodes to null nodes is akin to edges being {\it killed} or {\it birthed} along geodesics; and (iii) We demonstrate that the proposed \ac{SGM-net} establishes a new state-of-the-art benchmark across synthetic and real-world shape graph registration datasets while improving the inference time by {\bf an order of magnitude} over comparable baselines. 
 
\section{Related Works} \label{related work}

There are several ways to organize the extensive literature on registration methods. 
One criterion is to consider {\bf object type}. For instance, one can divide the object types into point clouds, traditional graphs, or shape graphs. (Some other articles~\cite{deng2009imagenet,sivic2003video} cater to other objects, such as images and videos but are less pertinent to the current discussion.) Another criterion is the choice of {\bf cost function} used in optimization for registration. The third criterion is the {\bf method for optimization}. Rather than discussing the previous works individually, we have organized the literature according to these criteria in Table 1.   

\tabcolsep .8 pt
\renewcommand{\arraystretch}{1.5}
\begin{table}[h]
\caption{Taxonomy of recent matching papers. $*$ denotes our modifications of existing methods to handle shapes.} 
\label{table-review}
\begin{center}
\resizebox{.9 \textwidth}{!}{%
\begin{tabular}{|c|c|c|c|}
\hline
 & \multicolumn{3}{c|}{\textbf{Data Structure}} \\ \hline
\textbf{Opt. Method} & \textit{Point Clouds} & \textit{Graph Matching} & \textit{Shape Graph} \\ \hline
\textit{Direct Optimization} & \cite{chen2022sc2,flamary2014optimal} &  \cite{bunke2000graph,motta2019vessel, Gallagher2006MatchingSA,yan-survey, sun2020survey,Ma2021-graph-similarity-survey,sewell1998branch,myers2000bayesian} & \cite{fgm}$^*$ \\ \hline
\textit{Learning Based} & \cite{chen2020graph, shen2021accurate, sun2022topology,puy2020flot, saleh2022bending,eisenberger2020deep,mei2021cotreg} & \cite{Yu2020-graph-matching, Wang2019-graph-matching,Wu2021-gnn-survey, Milan2017-graph-matching,Wang2018-graph-matching, Zhang2019-graph-matching, Zanfir2018-graph-matching, Nowak2018-graph-matching, Zhang2019-graph-matching-2, Derr2021-graph-matching, Xu2019-graph-matching, Sinkhorn1964-doubly-stochastic-matrices, Wang2020-graph-matching, Sarlin20-graph-matching, Jiang2022-graph-matching} & \cite{ngm}$^*$ \\ \hline
 & \multicolumn{3}{c|}{\textbf{Properties of Registration Solution}} \\ \hline
\textit{Matching Constraints} & None & Node-Node & Node-Node \& Elastic Edge-Edge \\ \hline
\multicolumn{1}{|c|}{\textit{Objective Function}} & MSE, OT & Node Features, OT & Elastic Shape Metric \\ \hline
\textit{Shape Geodesics} & No & No & Yes \\ \hline
\end{tabular} %
}
\end{center}
\end{table}

\noindent 
{\bf Object Type}: At a basic level, one can treat any shape graph as a {\bf point cloud} without any order, structure, or semantic labels. A large number of papers have studied the matching of point clouds using both classical and modern techniques. The limitation of this approach (in the current context) is the lack of structure -- the points can be matched to semantically different parts or in arbitrary ordering. A node may be registered with an interior point on edge and vice-versa! Such registration is difficult to interpret, especially in shape analysis. 
The next object type (increasing in structure) is {\bf traditional graphs} with Euclidean features. A list of some representative papers that study Euclidean graph matching is presented in Table~\ref{table-review}. Here the nodes are constrained to match the nodes. Still, usually, there is no consideration of edge points or shapes of edges in the matching problem. 
Finally, the object types of interest here are {\bf shape-graphs}. Here not only does one match nodes to nodes and edges to edges, but also involves the shapes of edges as integral to the matching problem (see Fig.~\ref{fig:examples_objects} bottom). Every point on the graph, either a node or an edge point, is accounted for in this dense matching. Our work aims to address the latter type of dense registration of shape graphs. 
\\

\noindent {\bf Cost Function}: 
Classical point-cloud matching techniques often use mean squared error (MSE), while many graph matching uses affinity matrices involving node features. Recent papers~\cite{shen2021accurate,saleh2022bending, chen2022sc2, motta2019vessel} have successfully utilized the optimal transport (OT) framework for graph matching. Our interest lies in using elastic shape metrics for matching edges since the registration is used in ensuing statistical shape analysis under that same metric. Although the OT methods provide good registration performance for their chosen criterion, they may perform poorly under the shape metrics. There is no existing paper that combines the OT framework with elastic shape metrics to register complex objects. Additionally, the OT framework for graph matching solely addresses the node registration problem and does not explicitly incorporate edge registration. 
\\

\noindent {\bf Optimization Method}: 
Historically, researchers have used a combination of gradients, relaxation techniques, and stochasticity to reach optimal solutions. Recent papers have involved deep neural networks in solving optimization problems. Initial efforts exploited \acp{RNN}~\cite{Milan2017-graph-matching}, \acp{GCN}~\cite{Wang2018-graph-matching,Xu2019-graph-matching}, \acp{GNN}~\cite{Zhang2019-graph-matching-2,Zanfir2018-graph-matching,Nowak2018-graph-matching} and adversarial networks~\cite{Derr2021-graph-matching}, but their operation was limited to the training graphs~\cite{Wang2018-graph-matching,Xu2019-graph-matching,Zhang2019-graph-matching-2,Derr2021-graph-matching}, to graphs of fixed size~\cite{Milan2017-graph-matching}, or treated node embedding extraction as a preprocessing stage separate from the matching~\cite{Zanfir2018-graph-matching,Nowak2018-graph-matching}. Zanfir~and~Sminchisescu~\cite{Zanfir2018-graph-matching} combined
feature extraction with a \ac{CNN} and graph matching with a \ac{GNN} in an end-to-end fashion so that the node embeddings could be optimized to facilitate the pairing process.
They also introduced the Sinkhorn algorithm~\cite{Sinkhorn1964-doubly-stochastic-matrices} for obtaining doubly stochastic similarity matrices in a differentiable manner, outlining a learning framework that motivated several follow-up works.
Subsequent advances involved incorporation of cross-graph convolutions~\cite{Wang2019-graph-matching,Wang2020-graph-matching,Sarlin20-graph-matching,Jiang2022-graph-matching,Yu2020-graph-matching}, a neighborhood consensus strategy via graph coloring~\cite{Fey2020-graph-matching}, and an extension to association graphs and hypergraphs~\cite{ngm}.

Even though these methods match graphs, they mostly relegate edges to a secondary role. Most of the current matching benchmarks are essentially node-matching tasks
and contain minimal to no-edge information. For this reason, edges are often artificially created, such as using complete graphs~\cite{Milan2017-graph-matching}, connecting nodes through proximity~\cite{Zhang2019-graph-matching} or Delaunay triangulation~\cite{Zanfir2018-graph-matching}, or learning optimal graph topology~\cite{Jiang2022-graph-matching}. Furthermore, these edges are frequently left unweighted and only serve as a means for message passing in \acp{GNN}. A few works exploit edge similarity, but edge embeddings are a mere combination of their endpoint embeddings~\cite{ngm,Yu2020-graph-matching,Zanfir2018-graph-matching}. Therefore, edges are seldom a source of new information. In contrast, we investigate the registration of shape graphs where the shapes of edges are the primary source of information.

Finally, we mention some papers that are directly relevant to our work. Motta et al.~\cite{motta2019vessel} study the problem of node-registration in retinal blood vessels (RBVs) as graphs using OT but ignore the shapes of blood vessels. Also, their results are demonstrated only for selectively paired data (of similar graphs) and not for arbitrary RBVs as we do. Saleh et al.~\cite{saleh2022bending} use the notion of shape graphs for registering points across human body surfaces. Their definition of shape graphs is different from ours because our data objects are different (2D surface versus network of blood vessels). 

In summary, our goal in this paper is to \underline{register shape graphs} under an \underline{elastic shape metric} using \underline{DNNs}. Later in the experiment sections, we will compare our approach with some past methods that register shape graphs or can be easily modified for this purpose. 

\begin{table}
\centering
\caption{Key Notations Used in This Paper}
\begin{tabular}{|c|m{2.5cm}|c|m{2.5cm}|} 
\hline
\textbf{Notation} & \textbf{Explanation} & \textbf{Notation} & \textbf{Explanation} \\ [0.5ex] 
\hline\hline
$G$ & Shape graph & $K_e$ & Edge-affinity matrix \\ 
\hline
$K_p$ & Node affinity matrix & $D$ & Node distance matrix \\
\hline
$A$ & Adjacency matrix & $C, F$ & Graph connectivity matrices~\cite{fgm} \\
\hline
$\rule{0pt}{2.2ex}\bar{A}$ & Unweighted adjacency matrix & $P$ & Node permutation matrix \\
\hline
$V^{(k)}, E^{(k)}$ & $k^{\text{th}}$ step node-affinity and edge-affinity embedding, respectively & $d_g$ & Composite distance between two shape graphs \\
\hline
\end{tabular}
\label{table:notations}
\end{table}

\section{Proposed Framework} \label{proposed method}
In this section, we present the proposed deep neural-network, \ac{SGM-net}, for registering pairs of shape graphs. To facilitate \ac{SGM-net}, we first introduce a mathematical representation for shape graphs and
the objective function used in registration. To help the reader with these notations, they are also summarized in Table~\ref{table:notations}.

\subsection{Background: Shape graph representation} \label{shape graph Representation}

To represent the shapes of individual curves that form the edges, we use the elastic shape analysis~\cite{FDA}. Let $\beta : [0, 1] \rightarrow \mathbb{R}^k$ ($k=2, 3$) be a continuous function, representing a parameterized curve. We define the square-root velocity function (SRVF)~\cite{FDA} of $\beta$ as: $q(t) = {\dot{\beta }(t)}/{\sqrt{|\dot{\beta} (t)|}}$. The shape of $\beta$ is mathematically represented by its orbit under the re-parametrization group: 
$[q] = \{(q\circ \gamma) \sqrt{\dot{\gamma}}| \gamma \in \Gamma\}$, where $\gamma : [0, 1] \rightarrow [0, 1]$ is a diffeomorphism and operator $\circ$ is function composition. 
The shape space of planar curves is given by ${\cal S} = \{[q]: q \in \mathbb{L}^2([0, 1], \mathbb{R}^k) \}$ and the associated shape metric is: $d_s ([q_1], [q_2]) = \inf_{\gamma}\|q_1 - (q_2\circ \gamma) \sqrt{\dot{\gamma}}\|$, where $\|\cdot\|$ is $\mathbb{L}^2$ norm. Note that the null edge (edge of zero length) ${\bf 0} \in {\cal S}$ is included in the representation space. 

With this representation of an edge, we turn our attention to a shape graph. 
We represent a shape graph by a pair of variables: an adjacency matrix to capture its edge attributes and a node matrix to represent its node attributes. If the graph has $n$ nodes, then its adjacency matrix is $A \in {\cal S}^{n \times n}$, where $A_{i,j}$ denotes the shape of edge starting from node $i$ and ending at node $j$, $i \neq j$. If there is no edge between any two nodes, then $A_{i,j} = {\bf 0}$. The diagonal elements of $A$ are also equal to ${\bf 0}$.
Since in our original data, the edges are not directed, we define another matrix $\tilde{A} \in {\cal S}^{n \times n}$, where $\tilde{A}_{ij}$ is simply a reflection re-parameterization of $A_{ij}$. When matching edges across graphs, we compare both $A_{ij}$ and $\tilde{A}_{ij}$ and keep the smallest distance of all combinations. 
 Let $u \in \mathbb{R}^{n \times k}$ denote the attributes of $n$ nodes. Each node is ascribed a vector attribute $u_i \in \real^k$. In this paper, we use $k=2$ with the planar coordinates of the nodes as their attributes, but other choices will work as well. Then, the pair $(A, u)$ defines a graph $G$. 

To reduce algorithmic complexity, we will use an alternative way to index edges in a graph. Let $s = 1, \dots, n_e$ and $s' = 1,\dotsm n'_e$ index the non-zero edges in graphs $G$ and $G'$, respectively. 
For a graph $G$, we also define two indicator matrices $C \in \mathbb{R}^{n \times n_e}$ and $F \in  \mathbb{R}^{n \times n_e}$ as follows. If there is an actual edge $A_{kl}$ originating from node $i$, then the entry $(i, n(k-1)+l)$ in $C$ is set to one, and zero otherwise. $F$ is the same, except it uses the edges in $\tilde{A}$ instead of $A$. $C$ and $F$ are called the {\it graph connectivity matrices}~\cite{fgm} and their dimensions are specified later.
\\

\noindent {\bf Shape Graph Metric and Registration}: 
Next, we define a metric for comparing two shape graphs, say $G = (A, u)$ and $G^{\prime} = (A^{\prime}, u^{\prime})$. At first, assume that the two graphs have the same number of nodes, say $n$. (We will generalize to different later in this section.) The number of edges in $G$ and $G'$ are $n_e$ and $n'_e$, respectively.  The edge metric $d_e$ is given by: 
$d_e(A_{i, j}, A^{\prime}_{i, j}) = \min\{ 
d_s (A_{i, j}, A^{\prime}_{i, j}), d_s (\tilde{A}_{i, j}, A^{\prime}_{i, j})\}$,
where $d_s$ is the shape metric defined above. Additionally, let the node distance matrix to be $D \in \real^{n \times n}$ such that $D_{ij} = \|u_i - u_j^{\prime}\|$, for any $u_i, u_j^{\prime} \in \real^2$, where $\|\cdot\|$ is standard vector norm. Finally, the composite distance between two graphs $G$ and $G^{\prime}$ becomes:
$$
d_g(G, G^{\prime}) =\lambda (\sum_{i, j} d_e(A_{i, j}, A^{\prime}_{i, j})) + (1-\lambda) \text{Tr}(D)\ ,
$$ 
where the relative weight parameter $0<\lambda <1$ controls the balance between the contributions of the nodes and edges.

Let ${\cal P}$ denote the set of all $n \times n$ permutation matrices. For any $P \in {\cal P}$, let $G*P$ represent the graph obtained by taking the nodes of $G$ and re-ordering them according to $P$. $ PAP^T$ denotes the new adjacency matrix (the matrix multiplication here implies changing the ordering of rows and columns of $A$, $\tilde{A}$), and $Pu$ stands for the new node attribute matrix. The graph registration problem is then given by: 
$$
\boxed{\hat{P} = \arg \min_{P \in {\cal P}} d_g(G, G^{\prime}*P)}\ .
$$
That is, we find a re-ordering of nodes of $G^{\prime}$ so that they are best matched with the nodes of $G$ to minimize $d_g$. 
\\

\noindent {\bf Building Affinity Matrices}: 
One can also formulate this problem using two large affinity matrices as follows. 
Define an edge-affinity matrix $K_e \in \mathbb{R}^{n_e \times n_e'}$ is given by: 
\begin{equation} \label{eq:Ke}
K_e^{s,s'} =
\lambda (1-\frac{d_e(A_{s}, A^{\prime}_{s'})}{\alpha}),\ \ \ \alpha = \max \{d_e(A_{s}, A^{\prime}_{s'})\}\ .
\end{equation}
An element $K_e^{s,s'}$  of this matrix compares the shape of edge $s$ of $G$ with the edge $s'$ of $G'$. The closer this value is to one, the more similar the two shapes are. 
Similarly, the node affinity matrix $K_p \in \mathbb{R}^{n \times n}$  is given by:
\begin{equation} \label{eqn:Kp}
K_p^{i,j} = (1-\lambda) (1 - \frac{D_{ij}}{\eta})\ \ \ , \ \eta = \max \{D_{ij}\}\ .
\end{equation}

\noindent {\bf Factorized formulation of QAP}: 
The pairwise graph-matching problem can be rewritten as a quadratic assignment problem (QAP) \cite{QAP} according to 
\begin{equation}
\max_{P \in {\cal P}} \ \text{vec}(P)^{\text{T}}K \text{vec}(P), \label{eq:FGM-original}
\end{equation}
where $K \in \real^{n^2 \times n^2}$ is called the {\it composite} affinity matrix between two graphs. Its diagonal and off-diagonal elements store the node and edge affinities across the graphs. Zhou~\etal~\cite{fgm} showed that $K$ can be factorized exactly as: 
$$
K = \text{diag}(\text{vec}(K_p)) + (C^{\prime}\otimes C)\text{diag}(\text{vec}(K_e))(F^{\prime}\otimes F)^{\text{T}},
$$
where $C, F$ and $C',F'$ are the graph connectivity matrices for $G$ and $G'$ and $\otimes$ denotes the Kronecker product. 
We note that $(C' \otimes C), (F'\otimes F)^T$ are both of size $(n^2 \times n_e n_e')$, $\text{diag}(\text{vec}(K_e))$ is of size $(n_en_e' \times n_en_e')$ and $\text{diag}(\text{vec}(K_p))$ is of size $(n^2 \times n^2)$.
Plugging the factorized formulation into QAP leads to an equivalent objective function:
$$
\max_{P \in {\cal P}} \{ \text{Tr}\left(K_p^T P \right) +
\text{Tr}\left(K_e^T (C^T P C^{\prime} \circ F^T P F^{\prime}) \right)\},
$$
where $\circ$ denotes the Hadamard product of matrices. 
\\

\noindent {\bf Using Null Nodes to Improve Registration}: 
In case the graphs $G, G'$ have different number of nodes, say $n, n^{\prime}$, where $n \leq n^{\prime}$, we can append $G$ with $m= n^{\prime} - n$ null nodes to bring them to the same size of $n^{\prime}$. Null nodes are artificially added to a graph with some chosen attributes to help balance the graph matching problem. One can either choose their attributes explicitly or choose the corresponding entries in the final affinity matrices. 
In this paper, we handle the null nodes by  
extending the node distance matrix $D$ to $\tilde{D}$ and node affinity matrix $K_p$ to $\tilde{K}_p$ as follows. Set
$\tilde{D} = \left[ \begin{array}{c} D \\ 
\epsilon \boldsymbol{1}_{m \times n^{\prime}}
\end{array} \right]$, where $\boldsymbol{1}_{(m \times n')}$ is an $(m \times n')$ matrix of ones. We use this $\tilde{D}$ to compute the larger node affinity matrix $\tilde{K}_p \in \mathbb{R}^{n' \times n'}$ as per Eqn.~\ref{eqn:Kp}.
Here $\epsilon= \text{Tr}(D) / n$ is chosen to penalize matching of real nodes in $G'$ with null nodes in $G$. We also extend the graph connectivity matrices $C, F\in \mathbb{R}^{n\times n_e}$ to $\tilde{C}, \tilde{F} \in \mathbb{R}^{n^{\prime} \times n_e}$ accordingly. The entries in the appended parts equal a constant $\xi$, which is a small positive number that allows taking logarithms later. 

\begin{figure*}
\centering
\includegraphics[width=0.89\linewidth]{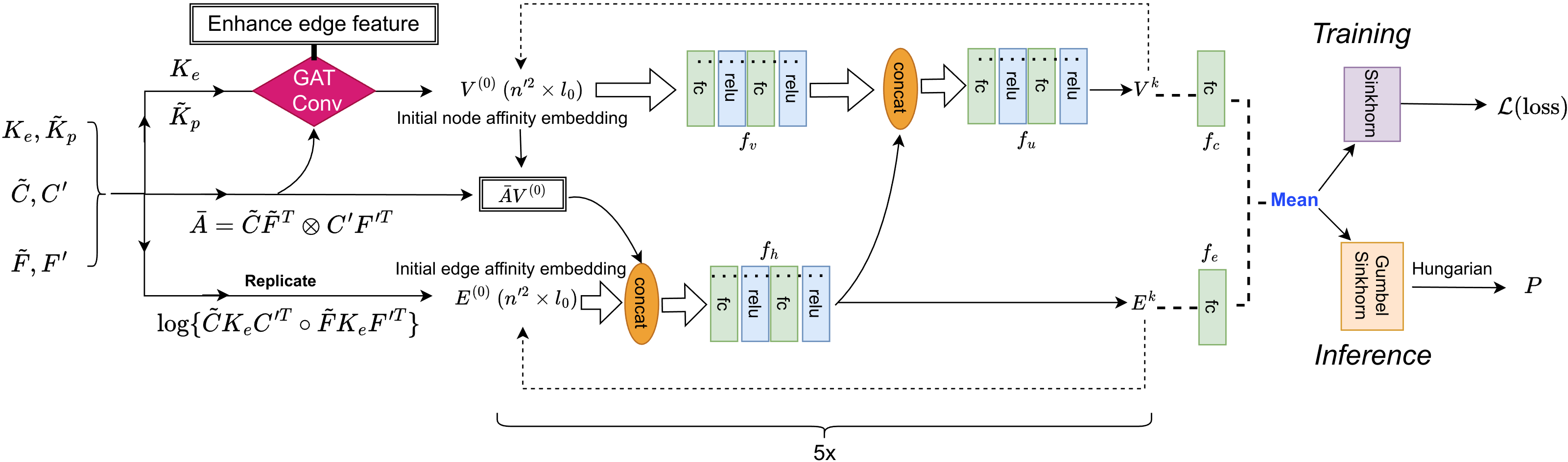} \\
\caption{The \textbf{overall architecture} for the proposed \ac{SGM-net} for pairwise graph matching is illustrated here. With an unsupervised loss function, This architecture uses two flows that discover and aggregate the node- and edge-affinity features separately. 
}
\label{fig:NGM solver}
\end{figure*}

\subsection{\ac{SGM-net} for shape graph matching}

Motivated by the factorized formulation of~\cite{fgm}, we propose a novel, deep network that solves the vertex classification using an association graph. However, instead of using a very large affinity matrix (as used in Eqn.~\ref{eq:FGM-original}), this network takes six smaller matrices: $ \tilde{K}_p \in \real^{n' \times n'}$, $K_e \in \real^{n_e \times n_e'}$, $\tilde{C}, \tilde{F} \in \real^{n' \times n_e}$, $C', F' \in \real^{n' \times n_e'}$ as inputs and outputs a node registration matrix. This node registration is then used to register points along the matched edges. 
The proposed network has two separate flows that discover and aggregate affinity features, one for nodes and one for edges. We shall call this network {\bf Shape-Graph Matching Network} (SGM-net). The overall architecture of the SGM-net is shown in Fig.~\ref{fig:NGM solver}, including both the training and inference procedures. We will provide more details in the following subsections. 

The proposed architecture provides several conceptual and practical advantages compared to a current SOTA. For instance, the \ac{NGM-net} of Wang~\etal~\cite{ngm} implements Lawler's QAP directly, using an association graph so that the graph-matching problem is equivalent to the vertex classification problem on this association graph. However, the \ac{NGM-net} builds the association graph directly from a computationally expensive (in space and time) composite affinity matrix ${K}$, which is of size $O(n^{\prime 2} \times n^{\prime 2})$. Thus, when applied to shape graphs, the \ac{NGM-net} can only handle graphs of a maximum of $60$ nodes or so on current hardware. Our approach uses smaller matrices and can handle much larger shape graphs. Besides, Wang~\etal used CNN layers to extract node features from natural images and constructed the edge features as differences between the corresponding node features. Thus, \ac{NGM-net} primarily registers the node features. Our network takes in the edge shapes as independent features and jointly performs registration using both edge and node features.  

\subsubsection{Matching aware embeddings of affinity matrices}\label{match embedding}
According to~\cite{fgm}, the matrices $\tilde{C}, \tilde{F}, C', F'$ can be re-interpreted as the node-edge connectivity matrices.
\\

\noindent {\bf Initial Edge Embedding}: Let the initial edge-affinity embedding be $E^{(0)}$, an $n'^2 \times l_0$ matrix, where each of the $l_0$ columns is a replicate of $\text{vec}(\log(\tilde{C} {K}_e C^{\prime T} \circ \tilde{F} {K}_e F^{\prime T}))$, $l_0$ is the initial affinity embedding size, $\circ$ is the Hadamard product, and $\log$ is elementwise. The intuition behind this equation is that if the edges in $G$ and $G^{\prime}$ are similar, then the corresponding connected nodes in the association graph tend to have a high classification score. 

\noindent {\bf Initial Node Embedding}:  We define an unweighted adjacency matrix $\bar{A} = \tilde{C} \tilde{F}^T \otimes C^{\prime} F ^{\prime T}$, which is an ${n^{\prime 2} \times n^{\prime 2}}$ sparse matrix of 1s and 0s. An entry in $\bar{A}$ indicates whether there is an edge between the corresponding two vertices in the association graph or not. 
We adapt the \ac{GATConv}~\cite{velivckovic2017graph} to construct the initial node embedding $V^{(0)} = \textit{GATConv}(\tilde{K}_p, \bar{A}, K_e) \in \mathbb{R}^{n^{\prime 2} \times l_0}$. The attention operator is applied to each node in the association graph, and the attention coefficients are computed based on $K_e$. This way different weights are assigned to each node, by aggregating the edge features from each node's neighbors. Readers are referred to the original paper~\cite{velivckovic2017graph} for details of \ac{GATConv}.  

The graph convolutional step updates node and edge embeddings iteratively according to:
\begin{eqnarray} \label{Mebb}
V^{(k)}_i &=& f_u\left(\left[f_h([E^{(k-1)} , \ \bar{A}V^{(k-1)}]_i),\  f_v(V^{(k-1)})\right]_i\right)\ ,
\\
\label{Mebb2}
E^{(k)}_i &=& f_h([E^{(k-1)},\ \bar{A}V^{(k-1)}]_i), \ \ i=1,2, \dots, n'^2,
\end{eqnarray}
where $V^{(k)}, E^{(k)}\in \mathbb{R}^{n^{\prime 2}\times l_k}$ are node- and edge-affinity embeddings. The notation $[\cdot \ , \  \cdot]$ implies a concatenation of matrices. The message passing functions $f_h, f_u: \mathbb{R}^{2l_{k-1}} \to \mathbb{R}^{l_{k}}$ and  $f_v:\mathbb{R}^{l_{k-1}} \to \mathbb{R}^{l_{k}}$ are all
implemented by networks with two fully-connected layers
and ReLU activation.

\subsubsection{Sinkhorn and Gumbel Sinkhorn network}
We utilize the Sinkhorn layer~\cite{sinkhorn-ranking,combmatch} to perform classification by turning the final embedding matrix $V^{(k)}$ and $E^{(k)}$, obtained in Eqn. \ref{Mebb2}, into a doubly stochastic matrix according to: for $i=1,2,\dots, n'^2$, we compute
$x_i = f_c(V^{(k)}_i) \in \real$, and $y_i = f_e(E^{(k)}_i) \in \real$, and set
$S= \exp (\frac{x+y}{2\tau}) \in \real^{n'^2 \times 1}$, where $\exp$ is elementwise.
Here $f_c, f_e: \mathbb{R}^{l_k} \rightarrow \mathbb{R}$ are both single fully-connected layer and $\tau$ is a normalization constant.
After reshaping the classification score matrix $S$ into $\mathbb{R}^{n^{\prime} \times n^{\prime}}$, the doubly-stochastic matrix can be obtained by element-wise division on rows and columns:
$
S = S \oslash (\boldsymbol{1}_{n^{\prime}}\boldsymbol{1}_{n^{\prime}}^{\text{T}} S),\ S = S \oslash (S \boldsymbol{1}_{n^{\prime}}\boldsymbol{1}_{n^{\prime}}^{\text{T}})$,
where symbol $\oslash$ is element-wise division. 

Similar to \cite{ngm}, the Gumbel Sinkhorn layer~\cite{gumbel-sinkhorn} is a post-selection technique that is applied at the inference stage. This layer enables searching over the space of all possible permutation matrices and uses the Hungarian algorithm~\cite{Hungarian} to select the one with the highest objective score. The Gumbel Sinkhorn layer is the same as the Sinkhorn layer, except that now $S$ is defined as $
 \exp (\frac{x+y+g}{2\tau})$, where $g$ is sampled from standard Gumbel distribution with CDF $e^{-e^{-x}}$. Although the Gumbel Sinkhorn layer significantly boosts the performance, it also adds to the computational cost. 
 A good balance between the performance and the computational cost can be achieved by cross-validation.

\subsubsection{Training Loss}
\label{Loss}
As shape graphs are constructed from real-world 3D images~\cite{bullitt2002volume,Stare,Drive}, there is no ground truth information available for training. During the training stage, the network performs optimization on the loss function 
$$
{\cal L}(S) = -\text{Tr}\left(\tilde{K}_p^T S \right) -
\text{Tr}\left(K_e^T (\tilde{C}^T S {C}^{\prime} \circ \tilde{F}^T S {F}^{\prime}) \right),
$$
where $\tilde{K}_p, K_e$ are the given affinity matrices, and $S$ is the output from the Sinkhorn layer. The objective score is the negative of the loss function. 

\subsection{Implementation Details}
Experiments are conducted on a Linux workstation with Nvidia RTX A6000 (48GB) GPU and AMD Ryzen 3900x CPU @ 4.6GHz with 64GB RAM. 
In all experiments reported here, we set the weight parameter $\lambda$ to be $0.5$, indicating equal importance for node and edge attributes. The training takes 120 epochs in total, the learning rate starts at $1 \cdot 10^{-5}$ and decays by half at $40^{\text{th}}$ and $80^{\text{th}}$ epoch, respectively. The hyperparameters of Sinkhorn and Gumbel Sinkhorn are chosen to be the same as the QAPLIB instances' setting~\cite{burkard1997qaplib} discussed in \cite{ngm}. The initial affinity-embedding size is $l_0 = 60$ and during the iterative graph convolution step, we use five graph convolution layers with channel sizes $l_1 = l_2 = l_3 = l_4 = l_5 = 256$. When we perform comparisons, the \ac{FGM}~\cite{fgm} is implemented in MATLAB while the \ac{SGM-net} has been implemented in Pytorch.

\begin{figure*}
\begin{center}
\begin{subfigure}{.48\textwidth}
\includegraphics[width=\linewidth]{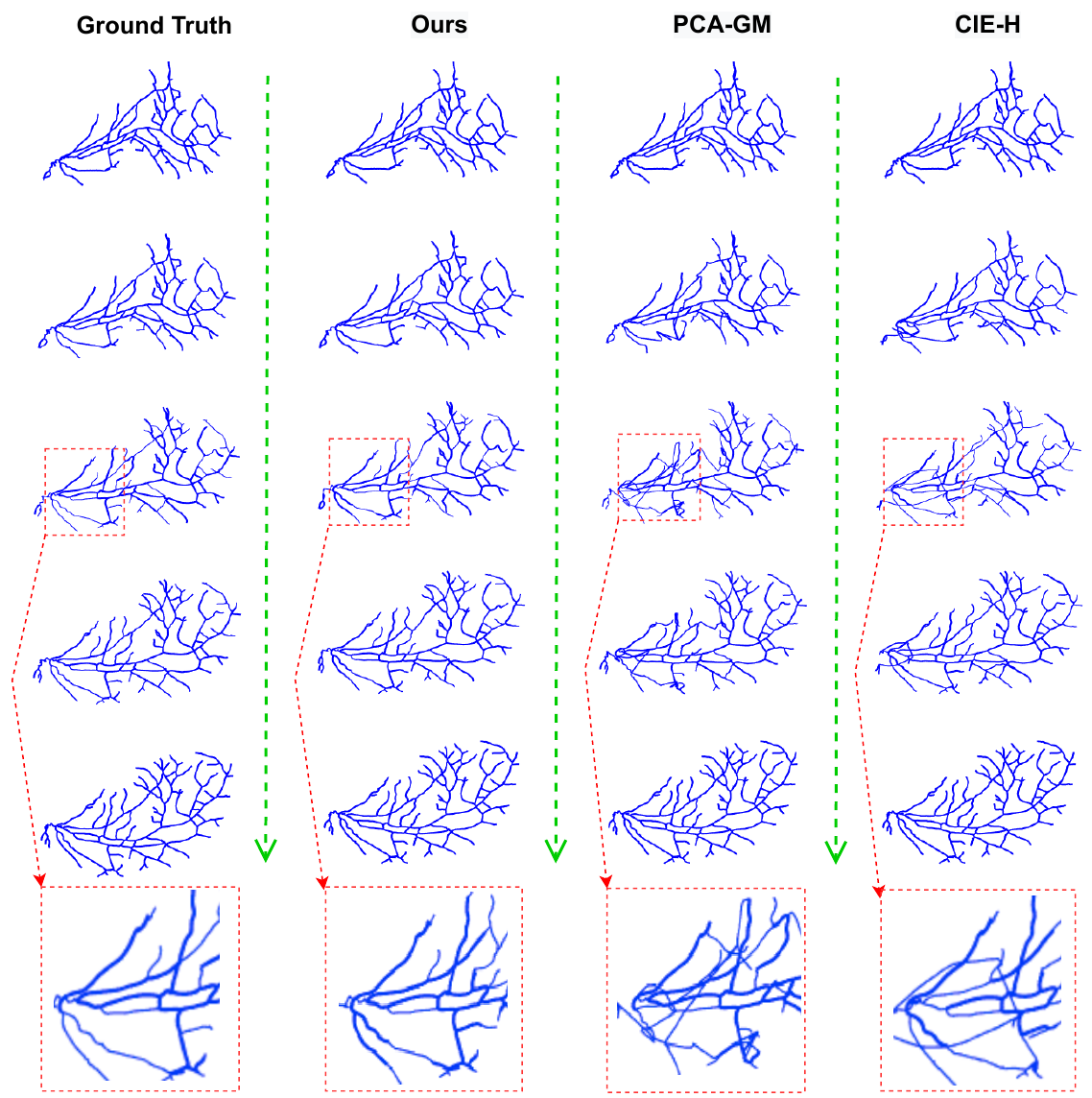}
   \caption{2D RBV networks}
   \label{fig: geodesics}
\end{subfigure}%
\qquad
\begin{subfigure}{.457\textwidth}
\includegraphics[width=\linewidth]{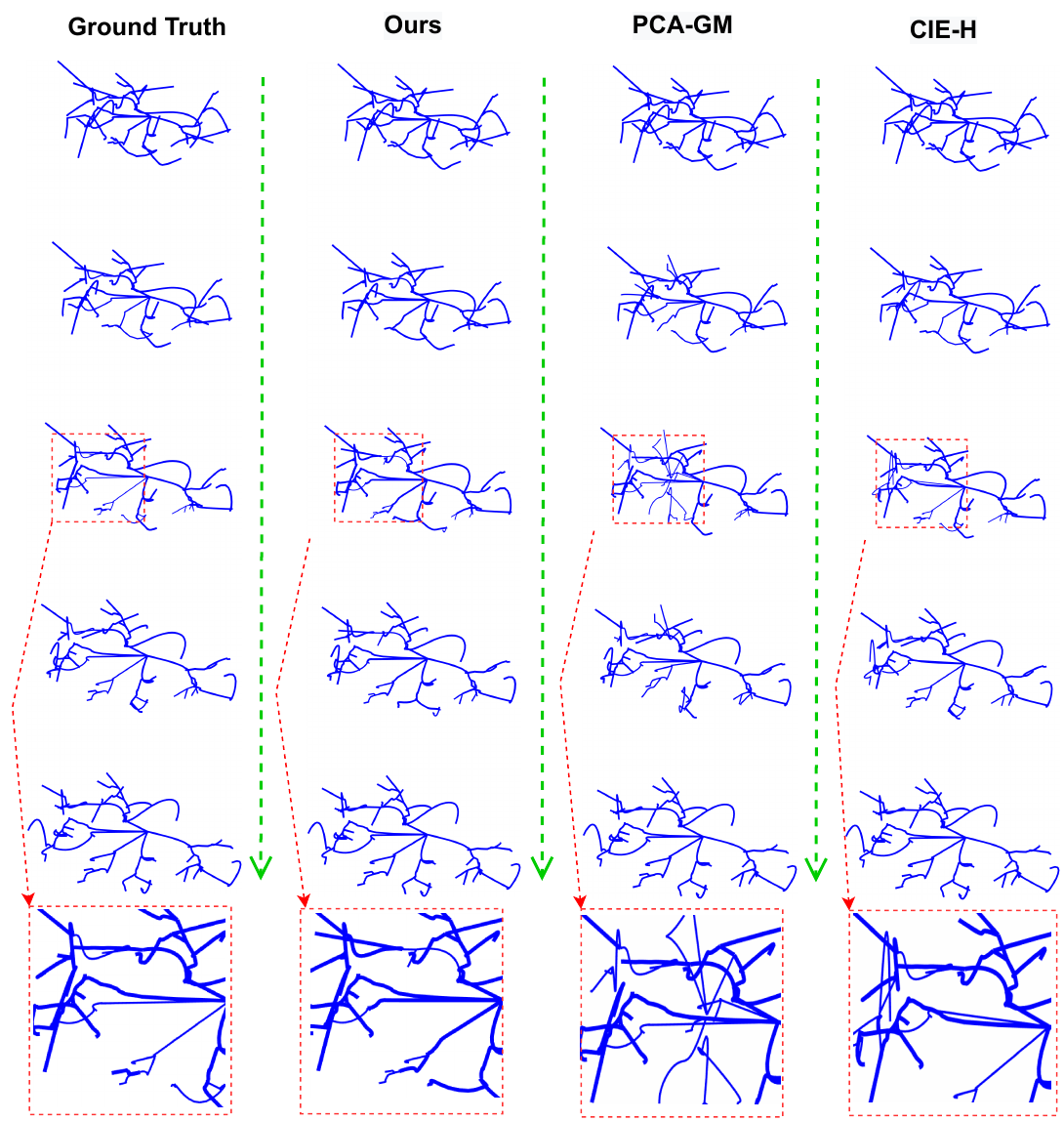}
\caption{3D Microglia dataset}
\label{fig: geo-neuro}
\end{subfigure}
\end{center}
\caption{{\bf Shape Geodesics}: Each column shows a geodesic between the top and the bottom shape graphs. The red dashed box at the bottom is a zoomed-in view of the selected area at the halfway point (3rd row). Our approach shows a natural deformation of blood vessels from one shape to another. NGM~\cite{ngm} failed to run due to its memory needs.}
\end{figure*}
\section{Experiments}\label{Experiments}
In this section, we provide results on the registration of shape graphs using the proposed SGM-net on both simulated and real datasets. The results obtained from SGM-net are compared with several SOTA methods including FGM and some recent neural networks designed for graph matching. In some cases where the original framework does not utilize edge-shape features, we have tried to incorporate and adapt so as to perform fair comparisons. 

\subsection{Data}
We use both synthetic and real-world datasets to evaluate our approach. A brief description of each dataset and its characteristics are presented below.
\\

\noindent
{\bf Real Dataset}: The first set is for 2D Retinal Blood Vessel (RBV) networks extracted from colored fundus images. The two well-known datasets~\cite{Stare,Drive} have around $240$ graphs with an average of $150$ nodes/graph (with $3000-5000$ total points on each graph: nodes plus edges). In order to study the matching and geodesic performance under different sizes,  we accordingly reduce their resolution. The second set is for 3D Brain Arterial Networks (BAN)~\cite{bullitt2002volume} with graphs having node sizes $120$ to $160$ (with $3500-4500$ total points per graph: nodes plus edges). The last set is for the 3D Microglia dataset~\cite{arshadi2021snt}, which is an open-source collection available on NeuroMorpho.Org~\cite{akram2018open}. The 3D Microglia dataset has around 90 graphs with an average of 100 nodes/graph (with 2500 - 300 total points on each graph: nodes plus edges). 
\\

\noindent
{\bf Synthetic Dataset}: 
The synthetic data is created by starting with actual 2D RBV networks and introducing noise, clutter, and distortions to generate paired shape graphs. We simulate three levels of distortion: low, medium, and high, between the paired graphs~\cite{BAN}.
A total of $900$ (registered) pairs ($300$ for each type) are simulated for training. The synthetic data for 3D BANs and Microglia datasets are simulated similarly. The size of nodes and total points for synthetic datasets are similar to the real datasets. 

For each experiment, we collect simulated and real data to form $1800$ pairs for training the network. The simulated data include samples from all three types (low, medium, and high levels of distortion). For testing, we include 
$54$ pairs ($18$ for each type) of synthetic graphs and $54$ pairs of real graphs. Note that for the real data, there is no ideal pairing available; we select graphs randomly to form pairs. The training and testing data have no overlap between them.

\subsection{Evaluation metrics.}
We evaluate registration methods on shape graphs of different complexities, both quantitatively and qualitatively (visually).
For quantitative results, we use the graph shape metric $d_g$ defined in Section \ref{shape graph Representation} to compare different registrations. Generally speaking, the smaller the distance, the better the registration. Additionally, we compute the
node objective score $\text{Tr}\left(\tilde{K}_p^T S \right)$ where a larger objective points to a better registration.
For qualitative results, we visualize the graph geodesics and compute sample means of several graphs under estimated registrations. A graph geodesic is a uniform speed of deformation path between the corresponding nodes and edges in $G$ and $G^{\prime}$. A sample mean is a representative graph capturing the common structures in the given graphs and is computed via an iterative approach. It registers the current mean estimate to all the graphs $G_i$, $i = 1,\cdots,m$ and averages the corresponding parts to update the mean $G_{\mu}$ until $\sum_{i=1}^m d_g({G_i, G_{\mu})^2}$ converges. For details, please refer to the paper~\cite{BAN}.

\subsection{State-of-the-art baselines} \label{peer}
As mentioned in Section 2, very few current DNNs perform shape graph registration under the shape metrics. We have adapted three recent DNNs for shape-graph matching to enable comparisons. These are: \textbf{1. PCA-GM}~\cite{Wang2019-graph-matching} approximates the distribution of node embeddings using cross-graph convolutions and then combines Sinkhorn's matrix scaling and binary cross-entropy to learn how to perform node assignment. We adapt this approach to also learn how to weigh neighbors in convolutions based on the shape of edges. 
 \textbf{2. CIE-H}~\cite{Yu2020-graph-matching} introduced a Hungarian Attention module that dynamically constructs a structured and sparsely connected layer, taking into account the most contributing matching pairs as hard attention during training. Here we only adopt the solver part from CIE-H, i.e., discard the image feature extraction part. \textbf{3. NGM}~\cite{ngm} learns with Lawler's QAP with given affinity matrices. In order to perform a fair comparison, here we modified the solver part from NGM so that it is entirely unsupervised with the goal of maximizing the same objective function as our method does. For a learning-free method, we also include a SOTA method \textbf{\ac{FGM}}~\cite{fgm} with  details as presented in~\cite{BAN}. 

\subsection{Evaluation results}
For synthetic experiments with known ground truth, we naturally compare results with the ground truth. For real data, there is no ground truth information, So we evaluate methods using the original objective function (the shape metric). Since PCA-GM and CIE-H require ground truth registration for training, we treat the results from \ac{FGM} as ground truth. \vspace*{-0.1in}\\

\noindent \textbf{Quantitative comparison.}
We compute the shape graph distance before the registration $d_g^B$ (i.e., the node order is arbitrary), and the graph distance after registration $d_g^R$. Then, the relative graph-distance reduction is obtained by $\frac{d_g^B - d_g^R}{d_g^B}$. The node objective score is defined in Section~\ref{Experiments}. Generally speaking, the larger the distance reduction and node objective score, the better the registration. 
Table~\ref{table:table performance 2D} and~\ref{table:table performance 3D}  show the comparison between our proposed \ac{SGM-net} and other recent graph-matching networks for 2D and 3D shape graphs, respectively. We can see that the \ac{SGM-net} outperforms other deep neural networks (CIE~\cite{Yu2020-graph-matching}, PCA-GM~\cite{Wang2019-graph-matching}, NGM\cite{ngm}) in all experiments in terms of graph distance reduction and node objective score. Besides, the SOTA QAP solver NGM failed to finish on the original Retinal Blood Vessel~\cite{Stare,Drive}, Brain Arterial Networks~\cite{bullitt2002volume} (graphs with around 150 nodes), and Microglia datasets~\cite{arshadi2021snt} (graphs with around 100 nodes) with an Nvidia RTX A100 (80GB) GPU. However, our method \ac{SGM-net} finishes the experiments on the original datasets with any 24GB GPU and in the meantime, the performance is guaranteed to be better. Between our \ac{SGM-net} and the SOTA learning-free method \ac{FGM}~\cite{fgm}, they achieve similar performances. For 2D RBV networks with 25 to 35 nodes, 45 to 65 nodes, and 3D BANs with 120 to 160 nodes, the \ac{SGM-net} slightly outperforms the \ac{FGM} in terms of graph distance reduction and node objective score. For 2D RBV networks with 120 to 200 nodes and 3D Micrglia dataset, \ac{FGM} slightly outperforms the \ac{SGM-net} in terms of graph distance reduction. However, the biggest justification for using \ac{SGM-net} comes in form of improved computational speed. \ac{SGM-net} incurs much lower cost as listed in Table~\ref{table:table performance 2D} and~\ref{table:table performance 3D}. We can see after the training stage,  the time gains for \ac{SGM-net} over FGM are substantial. \underline{\ac{SGM-net} is almost 220 times faster than \ac{FGM}}.
\\

\noindent \textbf{Geodesic deformations.}
We show some examples of geodesics obtained with different registrations from 2D RBV networks and 3D Microglia dataset in Fig.~\ref{fig: geodesics} and~\ref{fig: geo-neuro}, respectively. The results are drawn column-wise: in each column, we show graph $G$ (on top) and graph $G'$ (bottom) and the geodesic shapes in between them. Here we show only a few examples. To improve visual clarity, we prune\footnote{The entries in unweighted adjacency matrix which are smaller than $0.3$ are treated as $0$s so that the thin edges will not show in the display.} the intermediate graphs (along the geodesic) so that the overall structure of the graph stands out. The results in Fig.~\ref{fig: geodesics} and~\ref{fig: geo-neuro} suggest that 
the \ac{SGM-net} can achieve performance similar to that with ground truth registration. At the halfway point (the 3rd row), the ground truth and  \ac{SGM-net} registration largely preserve the salient structures and display a more natural deformation. However, The PCA-GM~\cite{Wang2019-graph-matching} and CIE-H~\cite{Yu2020-graph-matching} suffer from bad registration and provide distortion paths. The red dashed box in the bottom row highlights a selected region of the graphs at the halfway point (3rd row). We observe some crisscrossing and squeezing under PCA-GM and CIE-H registration, indicating bad registration. NGM failed to run the experiments due to large memory needs. 
 \vspace*{-0.1in} \\

\begin{figure*}
\begin{center}
\begin{subfigure}{.47\textwidth}
\includegraphics[width=\linewidth]{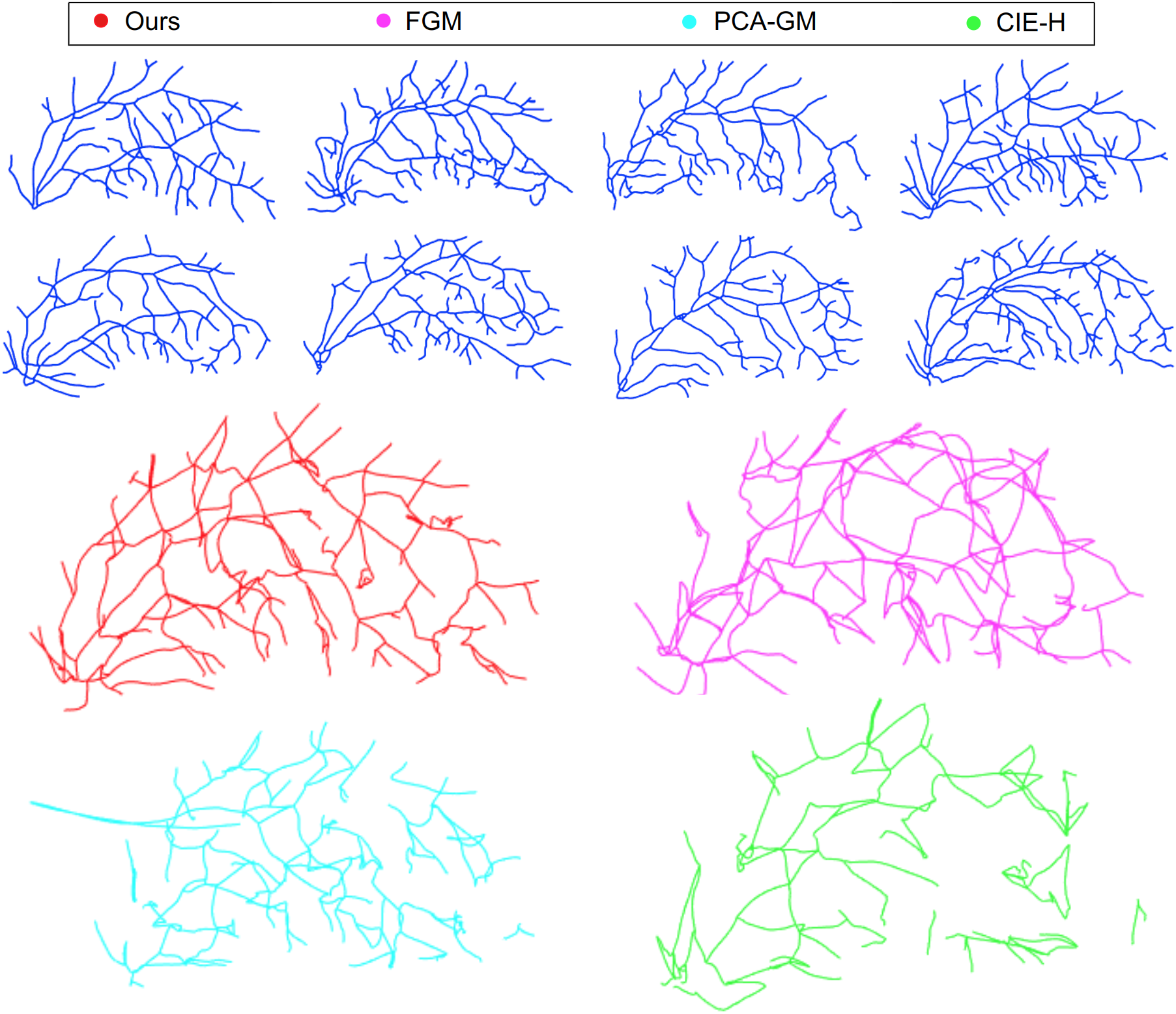}
   \caption{2D RBV networks}
   \label{fig: mean}
\end{subfigure}%
\qquad
\begin{subfigure}{.467\textwidth}
\includegraphics[width=\linewidth]{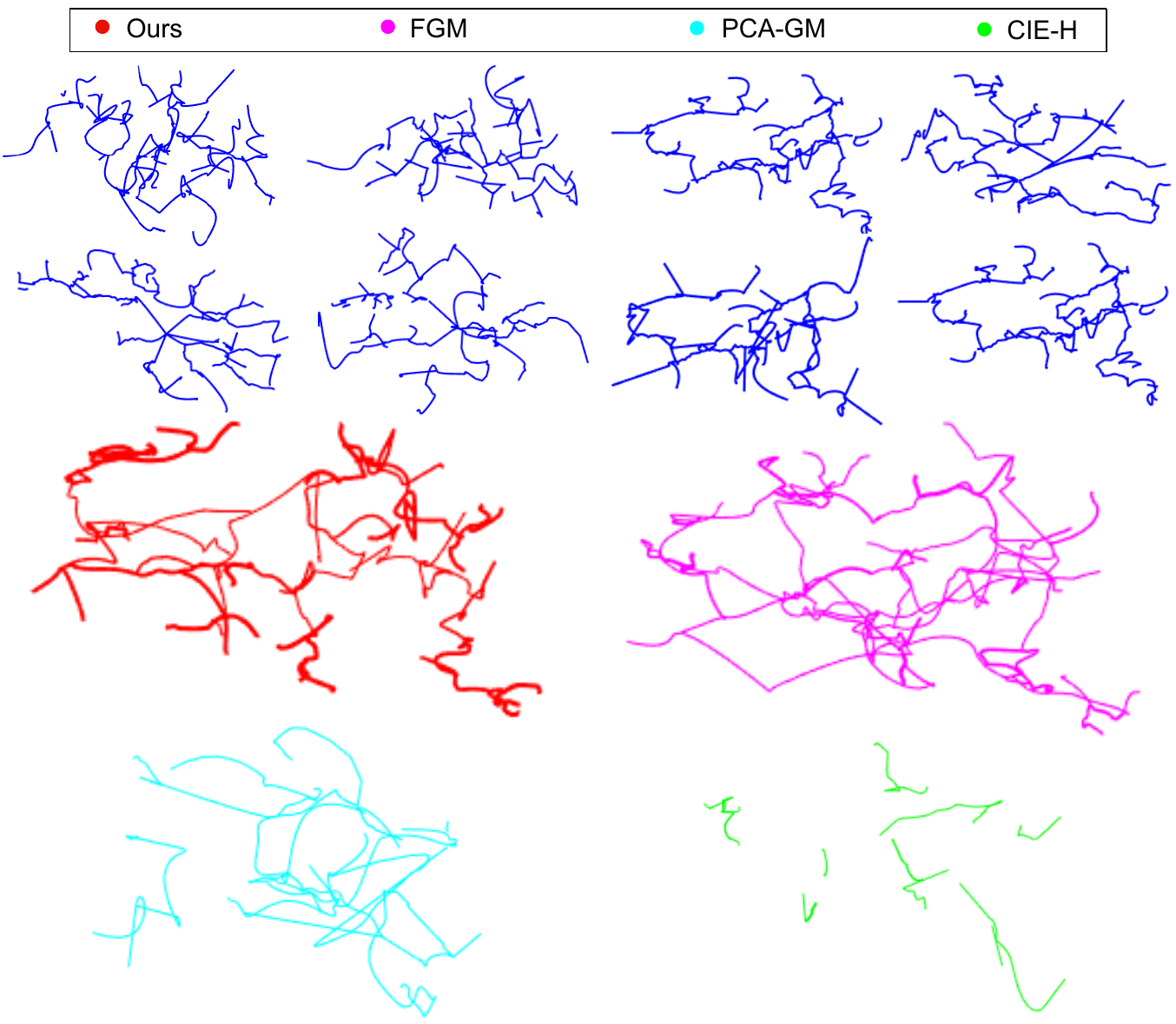}
\caption{3D Microglia dataset}
\label{fig: mean-neurom}
\end{subfigure}
\end{center}
\caption{Mean shapes of eight real shape graphs (blue) with node sizes $80 -120$ nodes (and $1500 - 3000$ total points), as computed by \ac{SGM-net} (red), \ac{FGM}~\cite{fgm} (magenta), PCA-GM~\cite{Wang2019-graph-matching} (cyan) and CIE-H~\cite{Yu2020-graph-matching} (green) registrations. NGM~\cite{ngm} failed to run the experiments due to its need for large memory.}
\end{figure*}

\noindent \textbf{Average shape graphs.}
Additionally, we compute averages of shapes of 2D RBV networks and 3D Microglia datasets under different registration methods and compare their results. We use the $8$ real shape graphs (with node sizes 80 to 160 nodes) shown in Fig.~\ref{fig: mean} and~\ref{fig: mean-neurom}. As earlier, we prune these mean graphs to improve visual clarity. 
As the results show, the two learning-based methods:  PCA-GM~\cite{Wang2019-graph-matching} and CIE-H~\cite{Yu2020-graph-matching}, fail to capture the overall structures in their averages. Specifically, several parts in the average shapes are disconnected from the main branches. 
The results for \ac{SGM-net} and \ac{FGM} appear similar, at least visually.  Similar to Table~\ref{table:table performance 2D} and~\ref{table:table performance 3D}, we see that the existing learning-based methods failed to handle the shape-based edge information, i.e., they hardly register the edges. Most of the edges in $G$ are registered to null edges in $G^{\prime}$. As a result, many edges will disappear after the pruning. NGM failed to finish the experiments due to its large memory requirement. 

Since the sample graphs are complex, it is difficult to easily evaluate the results visually. However, a closer inspection shows that the SGM-net better preserves the overall structures of these graphs. In summary, computing averages of shape graphs while preserving structure is a key challenge for any shape-graph matching method, as each iteration produces a new shape graph that other samples will be registered to, and the registration to this new shape graph is not learned during the training. This is another evidence that our method (\ac{SGM-net}) is robust and generalizes well.

\tabcolsep 1.9pt
\begin{table}
\centering
\caption{Comparisons of different registration methods on 2D RBV networks (For graph distance reduction and node scores the larger, the better). Node Size is the typical number of nodes and Graph Size is the total number of points, including nodes and edges, in each graph. The best results are in blue bold. For graphs with more than 80 nodes, \textbf{the NGM fails}.}
\label{table:table performance 2D}
\renewcommand{\arraystretch}{1.}
\begin{tabular}{ccccccccc}

\hline
\multicolumn{1}{|c}{}  &   & &\multicolumn{4}{c}{\textit{2D Retinal Blood Vessel Networks (RBVs)}}  &  & \multicolumn{1}{c|}{}  \\ 
\hline
\multicolumn{1}{|c}{\multirow{2}{*}{\rotatebox[origin=c]{90}{\scriptsize Type}}} & \multicolumn{2}{|c}{\multirow{1}{*}{\rotatebox[origin=c]{0}{\scriptsize Size}}} & \multicolumn{1}{|c|}{\multirow{2}{*}{Method}} & \multicolumn{1}{c|}{\multirow{1}{*}{Graph Dist.}} & \multicolumn{1}{c|}{\multirow{1}{*}{Node}} & \multicolumn{2}{c|}{Time} &  \multicolumn{1}{c|}{\scriptsize Speed Gain} \\ \cline{2-3}

\multicolumn{1}{|c}{} & \multicolumn{1}{|c}{\scriptsize Node} & \multicolumn{1}{|c}{\scriptsize Graph} & \multicolumn{1}{|c|}{}  & \multicolumn{1}{c|}{Reduc.(\%)} & \multicolumn{1}{c|}{Score} &  \multicolumn{2}{c|}{(s)}  &  \multicolumn{1}{c|}{\scriptsize over FGM}\\ 

\hline
\multicolumn{1}{|c}{\multirow{5}{*}{\rotatebox[origin=c]{90}{\scriptsize Real}}} & \multicolumn{1}{|c}{\multirow{5}{*}{\rotatebox[origin=c]{90}{\scriptsize 25-35}}} & \multicolumn{1}{|c}{\multirow{5}{*}{\rotatebox[origin=c]{90}{\scriptsize 625-950}}} & \multicolumn{1}{|c|}{CIE}    & \multicolumn{1}{c|}{36.21} &  \multicolumn{1}{c|}{12.92} & \multicolumn{2}{c|}{0.01} & \multicolumn{1}{c|}{}\\

\multicolumn{1}{|c}{} & \multicolumn{1}{|c}{} & \multicolumn{1}{|c}{} & \multicolumn{1}{|c|}{PCA-GM} & \multicolumn{1}{c|}{39.26} & \multicolumn{1}{c|}{13.19} & \multicolumn{2}{c|}{0.1} & \multicolumn{1}{c|}{}\\

\multicolumn{1}{|c}{} & \multicolumn{1}{|c}{} & \multicolumn{1}{|c}{} & \multicolumn{1}{|c|}{NGM} &  \multicolumn{1}{c|}{44.02} & \multicolumn{1}{c|}{13.41} & \multicolumn{2}{c|}{0.7} &\multicolumn{1}{c|}{} \\

\multicolumn{1}{|c}{} & \multicolumn{1}{|c}{} & \multicolumn{1}{|c}{} & \multicolumn{1}{|c|}{FGM} &  \multicolumn{1}{c|}{46.94} & \multicolumn{1}{c|}{13.06} & \multicolumn{2}{c|}{45} &\multicolumn{1}{c|}{} \\

\multicolumn{1}{|c}{} & \multicolumn{1}{|c}{} & \multicolumn{1}{|c}{} & \multicolumn{1}{|c|}{Ours}  & \multicolumn{1}{c|}{{\color[HTML]{0000FF} \textbf{47.01}}} & \multicolumn{1}{c|}{{\color[HTML]{0000FF} \textbf{13.43}}} & \multicolumn{2}{c|}{0.2} & \multicolumn{1}{c|}{225} \\ \hline \hline

\multicolumn{1}{|c}{\multirow{5}{*}{\rotatebox[origin=c]{90}{\scriptsize Real}}} & \multicolumn{1}{|c}{\multirow{5}{*}{\rotatebox[origin=c]{90}{\scriptsize 45-65}}} & \multicolumn{1}{|c}{\multirow{5}{*}{\rotatebox[origin=c]{90}{\scriptsize 1250-1850}}} & \multicolumn{1}{|c|}{CIE}  & \multicolumn{1}{c|}{39.34} & \multicolumn{1}{c|}{21.11} & \multicolumn{2}{c|}{0.07} & \multicolumn{1}{c|}{}\\

\multicolumn{1}{|c}{} & \multicolumn{1}{|c}{} & \multicolumn{1}{|c}{} &\multicolumn{1}{|c|}{PCA-GM}  & \multicolumn{1}{c|}{40.59}  & \multicolumn{1}{c|}{21.28} & \multicolumn{2}{c|}{0.63} & \multicolumn{1}{c|}{} \\

\multicolumn{1}{|c}{} & \multicolumn{1}{|c}{} & \multicolumn{1}{|c}{} &\multicolumn{1}{|c|}{NGM} & \multicolumn{1}{c|}{43.25} & \multicolumn{1}{c|}{ 21.30} & \multicolumn{2}{c|}{1.25} & \multicolumn{1}{c|}{} \\

\multicolumn{1}{|c}{} & \multicolumn{1}{|c}{} & \multicolumn{1}{|c}{} &\multicolumn{1}{|c|}{FGM} & \multicolumn{1}{c|}{45.58} & \multicolumn{1}{c|}{20.30} & \multicolumn{2}{c|}{167} & \multicolumn{1}{c|}{} \\

\multicolumn{1}{|c}{} & \multicolumn{1}{|c}{} &\multicolumn{1}{|c}{} & \multicolumn{1}{|c|}{Ours} & \multicolumn{1}{c|}{{\color[HTML]{0000FF} \textbf{47.20}}} & \multicolumn{1}{c|}{{\color[HTML]{0000FF} \textbf{21.31}}} & \multicolumn{2}{c|}{0.7} & \multicolumn{1}{c|}{240}  \\ \hline \hline

\multicolumn{1}{|c}{\multirow{4}{*}{\rotatebox[origin=c]{90}{\scriptsize Real}}} & \multicolumn{1}{|c}{\multirow{4}{*}{\rotatebox[origin=c]{90}{\scriptsize 120-200}}} & \multicolumn{1}{|c}{\multirow{4}{*}{\rotatebox[origin=c]{90}{\scriptsize 3500-4500}}} &\multicolumn{1}{|c|}{CIE} & \multicolumn{1}{c|}{32.54} & \multicolumn{1}{c|}{57.10} & \multicolumn{2}{c|}{0.11} & \multicolumn{1}{c|}{} \\

\multicolumn{1}{|c}{} & \multicolumn{1}{|c}{} &\multicolumn{1}{|c}{} & \multicolumn{1}{|c|}{PCA-GM} & \multicolumn{1}{c|}{34.24} & \multicolumn{1}{c|}{ 57.66} & \multicolumn{2}{c|}{3.31} & \multicolumn{1}{c|}{} \\

\multicolumn{1}{|c}{} & \multicolumn{1}{|c}{} &\multicolumn{1}{|c}{} & \multicolumn{1}{|c|}{FGM} & \multicolumn{1}{c|}{{\color[HTML]{0000FF} \textbf{44.72}}} & \multicolumn{1}{c|}{57.09} & \multicolumn{2}{c|}{1120} & \multicolumn{1}{c|}{} \\

\multicolumn{1}{|c}{} & \multicolumn{1}{|c}{} &\multicolumn{1}{|c}{} & \multicolumn{1}{|c|}{Ours} & \multicolumn{1}{c|}{42.83} & \multicolumn{1}{c|}{{\color[HTML]{0000FF} \textbf{59.27}}} & \multicolumn{2}{c|}{5.31} & \multicolumn{1}{c|}{211} \\ \hline \hline

\multicolumn{1}{|c}{\multirow{4}{*}{\rotatebox[origin=c]{90}{\scriptsize Synthetic}}} & \multicolumn{1}{|c}{\multirow{4}{*}{\rotatebox[origin=c]{90}{\scriptsize 120-200}}} & \multicolumn{1}{|c}{\multirow{4}{*}{\rotatebox[origin=c]{90}{\scriptsize 3500-5500}}} & \multicolumn{1}{|c|}{CIE} & \multicolumn{1}{c|}{45.23} & \multicolumn{1}{c|}{50.92} & \multicolumn{2}{c|}{0.09}& \multicolumn{1}{c|}{}  \\

\multicolumn{1}{|c}{} & \multicolumn{1}{|c}{} & \multicolumn{1}{|c}{} & \multicolumn{1}{|c|}{PCA-GM} & \multicolumn{1}{c|}{49.12} & \multicolumn{1}{c|}{49.99} & \multicolumn{2}{c|}{3.13}& \multicolumn{1}{c|}{}  \\

\multicolumn{1}{|c}{} & \multicolumn{1}{|c}{} & \multicolumn{1}{|c}{} & \multicolumn{1}{|c|}{Gnd.Truth} & \multicolumn{1}{c|}{50.09} & \multicolumn{1}{c|}{50.93} & \multicolumn{2}{c|}{--}& \multicolumn{1}{c|}{}  \\

\multicolumn{1}{|c}{} & \multicolumn{1}{|c}{} & \multicolumn{1}{|c}{} & \multicolumn{1}{|c|}{Ours} & \multicolumn{1}{c|}{{\color[HTML]{0000FF} \textbf{52.84}}} & \multicolumn{1}{c|}{{\color[HTML]{0000FF} \textbf{50.93}}} & \multicolumn{2}{c|}{5.18}& \multicolumn{1}{c|}{}  \\ 
\hline
\end{tabular}
\end{table}

\tabcolsep 1.9pt
\begin{table}
\centering
\caption{Comparisons of different registration methods on 3D BAN and Microglia datasets: (For graph distance reduction and node scores the larger, the better). Node Size is the typical number of nodes and Graph Size is the total number of points, including nodes and edges, in each graph. The best results are in blue bold. For graphs with more than 80 nodes, \textbf{the NGM fails}.}
\label{table:table performance 3D}
\renewcommand{\arraystretch}{1.}
\begin{tabular}{ccccccccc}

\hline
\multicolumn{1}{|c}{}  &   & &\multicolumn{4}{c}{\textit{3D Brain Arterial Networks (BANs)}}  &  & \multicolumn{1}{c|}{}  \\ 
\hline
\multicolumn{1}{|c}{\multirow{2}{*}{\rotatebox[origin=c]{90}{\scriptsize Type}}} & \multicolumn{2}{|c}{\multirow{1}{*}{\rotatebox[origin=c]{0}{\scriptsize Size}}} & \multicolumn{1}{|c|}{\multirow{2}{*}{Method}} & \multicolumn{1}{c|}{\multirow{1}{*}{Graph Dist.}} & \multicolumn{1}{c|}{\multirow{1}{*}{Node}} & \multicolumn{2}{c|}{Time} &  \multicolumn{1}{c|}{\scriptsize Speed Gain} \\ \cline{2-3}

\multicolumn{1}{|c}{} & \multicolumn{1}{|c}{\scriptsize Node} & \multicolumn{1}{|c}{\scriptsize Graph} & \multicolumn{1}{|c|}{}  & \multicolumn{1}{c|}{Reduc.(\%)} & \multicolumn{1}{c|}{Score} &  \multicolumn{2}{c|}{(s)}  &  \multicolumn{1}{c|}{\scriptsize over FGM}\\ 

\hline
\multicolumn{1}{|c}{\multirow{4}{*}{\rotatebox[origin=c]{90}{\scriptsize Synthetic}}} & \multicolumn{1}{|c}{\multirow{4}{*}{\rotatebox[origin=c]{90}{\scriptsize 120-160}}} & \multicolumn{1}{|c}{\multirow{4}{*}{\rotatebox[origin=c]{90}{\scriptsize 3500-4500 }}} & \multicolumn{1}{|c|}{CIE} & \multicolumn{1}{c|}{53.53} & \multicolumn{1}{c|}{28.87} & \multicolumn{2}{c|}{0.07}& \multicolumn{1}{c|}{}  \\

\multicolumn{1}{|c}{} & \multicolumn{1}{|c}{} & \multicolumn{1}{|c}{} & \multicolumn{1}{|c|}{PCA-GM} & \multicolumn{1}{c|}{58.52} & \multicolumn{1}{c|}{30.04} & \multicolumn{2}{c|}{1.28}& \multicolumn{1}{c|}{}  \\

\multicolumn{1}{|c}{} & \multicolumn{1}{|c}{} & \multicolumn{1}{|c}{} & \multicolumn{1}{|c|}{Gnd.Truth} & \multicolumn{1}{c|}{60.56} & \multicolumn{1}{c|}{{\color[HTML]{0000FF} {\textbf{31.07}}}} & \multicolumn{2}{c|}{--} & \multicolumn{1}{c|}{} \\ 

\multicolumn{1}{|c}{} & \multicolumn{1}{|c}{} & \multicolumn{1}{|c}{} & \multicolumn{1}{|c|}{Ours} & \multicolumn{1}{c|}{{\color[HTML]{0000FF} \textbf{61.71}}} & \multicolumn{1}{c|}{31.01} & \multicolumn{2}{c|}{2.71} & \multicolumn{1}{c|}{} \\ \hline

\multicolumn{1}{|c}{\multirow{4}{*}{\rotatebox[origin=c]{90}{\scriptsize Real}}} & \multicolumn{1}{|c}{\multirow{4}{*}{\rotatebox[origin=c]{90}{\scriptsize 120-160}}} & \multicolumn{1}{|c}{\multirow{4}{*}{\rotatebox[origin=c]{90}{\scriptsize 3500-4500 }}} & \multicolumn{1}{|c|}{\multirow{1}{*}{CIE}} & \multicolumn{1}{c|}{\multirow{1}{*}{21.67}} & \multicolumn{1}{c|}{\multirow{1}{*}{43.42}} & \multicolumn{2}{c|}{\multirow{1}{*}{0.1}} & \multicolumn{1}{c|}{} \\

\multicolumn{1}{|c}{} & \multicolumn{1}{|c}{} & \multicolumn{1}{|c}{} & \multicolumn{1}{|c|}{PCA-GM} & \multicolumn{1}{c|}{18.06} & \multicolumn{1}{c|}{42.07} & \multicolumn{2}{c|}{1.41}& \multicolumn{1}{c|}{}  \\

\multicolumn{1}{|c}{} & \multicolumn{1}{|c}{} & \multicolumn{1}{|c}{} & \multicolumn{1}{|c|}{FGM} & \multicolumn{1}{c|}{\multirow{1}{*}{34.89}} & \multicolumn{1}{c|}{45.23} & \multicolumn{2}{c|}{1114}& \multicolumn{1}{c|}{}  \\

\multicolumn{1}{|c}{\multirow{1}{*}{}} &\multicolumn{1}{|c}{} & \multicolumn{1}{|c}{\multirow{1}{*}{}} & \multicolumn{1}{|c|}{\multirow{1}{*}{Ours}} & \multicolumn{1}{c|}{{\multirow{1}{*}{\color[HTML]{0000FF} {\textbf{34.91}}}}} & \multicolumn{1}{c|}{{\multirow{1}{*}{\color[HTML]{0000FF} \textbf{46.17}}}} & \multicolumn{2}{c|}{\multirow{1}{*}{3.82}} & \multicolumn{1}{c|}{292} \\

\hline
\multicolumn{1}{|c}{}  &  \multicolumn{1}{|c}{} &\multicolumn{1}{|c}{}& \multicolumn{4}{c}{\textit{3D Microglia dataset}} & & \multicolumn{1}{c|}{}  \\ 
\hline
\multicolumn{1}{|c}{\multirow{4}{*}{\rotatebox[origin=c]{90}{\scriptsize Synthetic}}} & \multicolumn{1}{|c}{\multirow{4}{*}{\rotatebox[origin=c]{90}{\scriptsize 80-120}}} & \multicolumn{1}{|c}{\multirow{4}{*}{\rotatebox[origin=c]{90}{\scriptsize 1000-2000}}} & \multicolumn{1}{|c|}{CIE} & \multicolumn{1}{c|}{53.43} & \multicolumn{1}{c|}{33.31} & \multicolumn{2}{c|}{0.09}& \multicolumn{1}{c|}{}  \\

\multicolumn{1}{|c}{} & \multicolumn{1}{|c}{} & \multicolumn{1}{|c}{} & \multicolumn{1}{|c|}{PCA-GM} & \multicolumn{1}{c|}{53.31} & \multicolumn{1}{c|}{33.31} & \multicolumn{2}{c|}{0.21}& \multicolumn{1}{c|}{}  \\

\multicolumn{1}{|c}{} & \multicolumn{1}{|c}{} & \multicolumn{1}{|c}{} & \multicolumn{1}{|c|}{Gnd.Truth} & \multicolumn{1}{c|}{53.63} & \multicolumn{1}{c|}{{33.33}} & \multicolumn{2}{c|}{--} & \multicolumn{1}{c|}{} \\ 

\multicolumn{1}{|c}{} & \multicolumn{1}{|c}{} & \multicolumn{1}{|c}{} & \multicolumn{1}{|c|}{Ours} & \multicolumn{1}{c|}{{\color[HTML]{0000FF} \textbf{55.26}}} & \multicolumn{1}{c|}{\color[HTML]{0000FF} {\textbf{33.34}}} & \multicolumn{2}{c|}{.0.98} & \multicolumn{1}{c|}{} \\ \hline

\multicolumn{1}{|c}{\multirow{4}{*}{\rotatebox[origin=c]{90}{\scriptsize Real}}} & \multicolumn{1}{|c}{\multirow{4}{*}{\rotatebox[origin=c]{90}{\scriptsize 80-120}}} & \multicolumn{1}{|c}{\multirow{4}{*}{\rotatebox[origin=c]{90}{\scriptsize 1000-2000 }}} & \multicolumn{1}{|c|}{\multirow{1}{*}{CIE}} & \multicolumn{1}{c|}{ {{25.95}}} & \multicolumn{1}{c|}{\multirow{1}{*}{29.41}} & \multicolumn{2}{c|}{\multirow{1}{*}{0.08}} & \multicolumn{1}{c|}{} \\

\multicolumn{1}{|c}{} & \multicolumn{1}{|c}{} & \multicolumn{1}{|c}{} & \multicolumn{1}{|c|}{PCA-GM} & \multicolumn{1}{c|}{19.05} & \multicolumn{1}{c|}{28.52} & \multicolumn{2}{c|}{0.22}& \multicolumn{1}{c|}{}  \\

\multicolumn{1}{|c}{} & \multicolumn{1}{|c}{} & \multicolumn{1}{|c}{} & \multicolumn{1}{|c|}{FGM} & \multicolumn{1}{c|}{\multirow{1}{*}{\color[HTML]{0000FF} {\textbf{{35.62}}}}} & \multicolumn{1}{c|}{30.02} & \multicolumn{2}{c|}{290}& \multicolumn{1}{c|}{}  \\

\multicolumn{1}{|c}{\multirow{1}{*}{}} &\multicolumn{1}{|c}{} & \multicolumn{1}{|c}{\multirow{1}{*}{}} & \multicolumn{1}{|c|}{\multirow{1}{*}{Ours}} & \multicolumn{1}{c|}{{\multirow{1}{*}{34.93}}} & \multicolumn{1}{c|}{{\multirow{1}{*}{\color[HTML]{0000FF} \textbf{31.08}}}} & \multicolumn{2}{c|}{\multirow{1}{*}{1.26}} & \multicolumn{1}{c|}{230} \\ 

\hline
\end{tabular}
\end{table}

\subsection{Ablation study on synthetic dataset}
In this section, we evaluate the impact of our \ac{SGM-net}'s primary components by sequentially introducing them to a basic baseline model. We carry out these experimental modifications on 2D synthetic RBV networks with nodes ranging from 45 to 65. The results are presented in Table~\ref{table:abl}.

The basic baseline model only consists of a node affinity $K_p$ aggregation network, excluding the layers required for the aggregation of edge affinity $K_e$. 
We also explore the functionalities of the unsupervised loss and Gumbel sinkhorn layer by substituting them with a cross-entropy loss between the predicted and ground truth registrations, and a standard sinkhorn layer, respectively.
Additionally, we investigate the potential influence of hypothetical null nodes. Specifically, we consider two shape graphs, $G$ and $G^{\prime}$, having $n$ and $n^{\prime}$ nodes, where $n$ is less than or equal to $n^{\prime}$. The third row of Table~\ref{table:abl} (Without Null Nodes) represents a scenario where null nodes are not added to $G$. During the computation of evaluation metrics, we assume that null nodes in $G$ arbitrarily align with the unregistered nodes in $G^{\prime}$.

As demonstrated in the table, we observe a gradual improvement in performance as we systematically incorporate each component into our model. 

\begin{table}
\centering
\caption{Results of ablation studies, demonstrating the performance impact of various model components. Each row represents a variant of our model (SGM-net), where we progressively add components from top to bottom. The columns show the percentage reduction in graph distance and the node and edge scores for each variant. As can be observed, the complete model has the best performance.}
\label{table:abl}
\begin{tabular}{@{}llll@{}}
\toprule
 & \begin{tabular}[c]{@{}l@{}}Graph Dist.\\ Reduc. (\%)\end{tabular} & \begin{tabular}[c]{@{}l@{}}Node\\ Score\end{tabular} & \begin{tabular}[c]{@{}l@{}}Edge\\ Score\end{tabular} \\ \midrule
Baseline Node Affinity & 13.03 & 44.38 & 5.01  \\ \midrule
\begin{tabular}[c]{@{}l@{}}Node- + Edge-Affinity \\with Standard Sinkhorn, \\Cross-Entropy Loss\end{tabular} & 35.24 & 57.29 & 57.97 \\ \midrule
Without Null Nodes & 44.59 & 60.53 & 68.28 \\ \midrule
Complete Model (SGM-net) & 46.79 & 63.24 & 68.29 \\ 
\bottomrule
\end{tabular}
\end{table}

\section{Conclusion and Discussion} \label{Discussion}
This paper proposes a deep learning-based registration framework for handling complex geometrical structures -- shape graphs -- that appear in biological and anatomical data. This method uses unsupervised training and learning-based registration to provide state-of-the-art results and order-of-magnitude improvements in computational costs. The key distinction from the past methods is that it incorporates the shape metrics for curves in comparing edges across graphs. It obtains speed gains of $\approx 200-300$ times the SOTA technique FGM. The registration is performed under the shape graph metric and this efficient registration enables us to further develop tools for statistical analysis of shape graphs, including computations of geodesics and shape summaries.





\bibliography{main}
\bibliographystyle{IEEEtran}


\end{document}